\documentclass[preprint]{elsarticle}
\makeatletter
\def\ps@pprintTitle{%
 \let\@oddhead\@empty
 \let\@evenhead\@empty
 \def\@oddfoot{\centerline{\thepage}}%
 \let\@evenfoot\@oddfoot}
\makeatother

\usepackage{hyperref}

\usepackage{rotating}
\usepackage{float}
\usepackage{times}
\usepackage{epsfig}
\usepackage{graphicx}
\usepackage{amsmath}
\usepackage{amssymb}
\usepackage{tabularx}
\usepackage{caption}
\usepackage{multirow}

\def\eq{\texttt{=}}

\def\etal{\emph{et al.}}

\begin{document}

\begin{frontmatter}

\title{Egocentric View Hand Action Recognition by Leveraging Hand Surface and Hand Grasp Type}


\author[main]{Sangpil Kim}
\author[main]{Jihyun Bae}
\address[main]{Korea University, 145 Anam-ro, Seoul, South Korea}

\author[mymainaddress]{Hyunggun Chi}

\author[KETI]{Sunghee Hong}
\address[KETI]{Korea Electric Technology Institute,   Seongnam-si, Gyeonggi-do 13509, South Korea}

\author[DigiCAP]{Byoung Soo Koh}
\address[DigiCAP]{DigiCAP Inc., Gangseo-gu, Seoul, Korea}

\author[mymainaddress]{Karthik Ramani}


\address[mymainaddress]{Purdue University, West Laffayette, Indiana, USA}

\begin{abstract}
We introduce a multi-stage framework that uses mean curvature on a hand surface and focuses on learning interaction between hand and object by analyzing hand grasp type for hand action recognition in egocentric videos. The proposed method does not require 3D information of objects including 6D object poses which are difficult to annotate for learning an object's behavior while it interacts with hands. 
Instead, the framework synthesizes the mean curvature of the hand mesh model to encode the hand surface geometry in 3D space.
Additionally, our method learns the hand grasp type which is highly correlated with the hand action.
From our experiment, we notice that using hand grasp type and mean curvature of hand increases the performance of the hand action recognition. 
\end{abstract}

\begin{keyword}
\texttt{machine perception} \sep \texttt{hand action recognition} \sep \texttt{deep learning} 
\sep \texttt{surface modality}
\end{keyword}

\end{frontmatter}

\begin{figure}[h]
\begin{center}
    \centering
    \includegraphics[width=1.\textwidth,height=5cm]{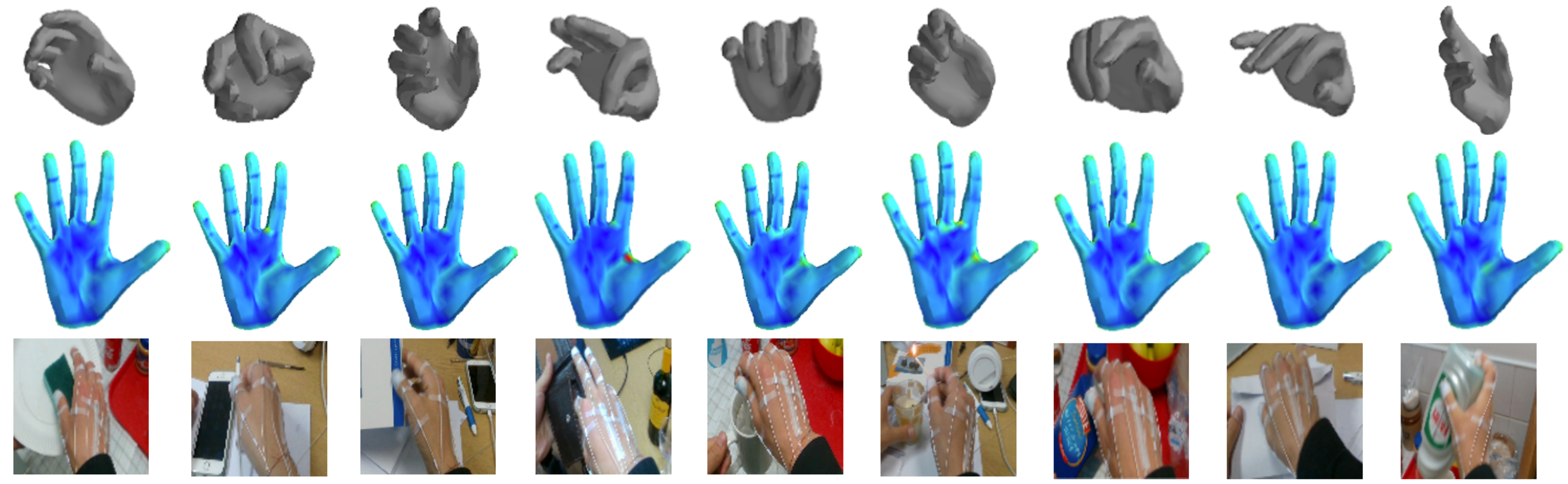}
    \caption{The first row shows fitted hand mesh models given the cropped hand image. Mean curvature distribution on hand mesh is shown in the second row. The values are color-coded from low value as blue to high value as red. The images on the last row are cropped hands from the egocentric videos.}
    \label{fig:curv_main}
\end{center}
\end{figure}

\section{Introduction}
People have been using hands in their daily activities to interact with objects and communicate with others.
Therefore, a significant amount of research has focused on hand action recognition~\cite{damen2018scaling,fathi2011learning,garcia2018first,li2018eye}.
Unlike hand gestures, hand actions involve objects, and therefore, understanding the grasp type of the hand is essential.
Also, the hand surface is dependent on what the hand is performing during the hand action.
This brings us to hand type and surface and their interaction with objects for hand action recognition.

The advancement of low-cost wearable sensors and augmented reality/virtual reality technologies motivates the computer vision community to tackle egocentric view hand action recognition. However, hand action recognition is extremely challenging because the perception model needs to be robust on diverse topology of both rigid and non-rigid objects. 
In cluttered real-world scenes, hand dorsum and objects create a large occlusion region of hands. 
Fingertips in particular, which are the major functional parts of the hands, are occluded in the egocentric view in many cases.
Additionally, the fast movement of a camera view makes the hand action prediction challenging~\cite{li2015delving}.
To address these problems, the computer vision community incorporates depth sensors~\cite{tang2017action,garcia2018first} and multiple cameras with different views to reduce occluded area of the hand regions~\cite{zimmermann2019freihand}.
Also, many works~\cite{tekin2019h+,sridhar2016real} have shown that considering hand and object interactions increases the accuracy of prediction of hand actions and 3D hand poses.

However, these works contain limitations as the following.
First, the works which use 6D pose of objects for learning hand-object interactions do not consider the complexity of object space and the cost of annotating 6D pose of objects.
The creation of the ground truth of 6D pose of objects often requires 3D mesh models since objects are occluded by the hand.
Additionally, there is a nearly infinite number of different objects with distinct topology and geometry.
Second, depth sensors are not robust under high luminance nature lights and are more expensive than RGB cameras. 
Third, there are no methods that focus on the relation of the hand surface geometry properties and the hand action recognition.
Lastly, multi-camera setting is also not efficient since it requires camera calibration and synchronising frames from multiple cameras is extremely difficult.

Our proposed method tackles these problems without using 6D pose of objects neither 3D information of the scene for the hand action recognition task.
In our pipeline, the grasp type is used as mid-level representations of hand-object interaction for identifying hand actions and mean curvature that represents the hand surface works as low-level representations which are geometrically descriptive~(see Fig.~\ref{fig:curv_main}).
These low-level and mid-level features of the hand are further transformed into high-level concept which is the hand action in the framework.
In biomechanics perspective, the hand action involves interaction with objects which is positively related to the grasp or pre-grasp types, the surface of the hand, and geometry of the object.
Additionally, we notice that the interactions between the primary and the assistive hand/object are important factors to learn hand-object interactions for the hand action recognition.

Our contributions are listed as follows:

\begin{itemize}
 \item We present a novel hand action estimation pipeline that learns the distribution of mean curvatures of the hand surface which imposes detailed geometric information. Since mean curvature is only dependent on the surface of the hand, the proposed pipeline tends to be location-invariant from the global position of each hand joint while interacting with an object. 

 \item We propose a learning methodology of hand-object interaction by identifying hand grasp types in the videos with deep neural networks, without using 3D information of neither the object nor hand for the hand action recognition task in egocentric-view videos.

 \item We constructed a taxonomy of hand grasp types that includes diverse hand grasp types while interacting with objects. 
With the taxonomy, we labeled the hand grasp types on every frame in the public video dataset and show their usage in hand action recognition task.

 \item We show that a hand action is not defined by a single hand grasp type but by combinations of diverse hand grasp types.
 With this finding, we predict the hand grasp type per frame and integrate them with temporal neural networks, which increases the performance of the hand action estimation. 
\end{itemize}

Each of the above contributions improves overall hand action recognition performance, and we show the improvement quantitatively in Section~\ref{sec:ex_ha}.
Our method outperforms the state-of-the-art methods without using 6D object pose, 3D hand joint locations, and depth information in the sophisticated hand action dataset.

\section{Related Work}

\textbf{Hand action recognition.}
Hand action is an activity that involves hand-object interaction.
Therefore, the hand pose is an important feature for hand action recognition~\cite{garcia2018first,tekin2019h+}. 
This is because the hand pose implies the object geometry and the grasp type, which is positively correlated with the hand action.
Hand pose can be accurately estimated by using a depth sensor~\cite{garcia2018first} since a depth image is texture-invariant and imposes 3D information, which is critical for estimating 3D hand poses.
Knowing the depth of scenes and the hand enhances the accuracy of the hand action estimator.
Hand action often involves object manipulation with the hand or with a tool. 
Therefore, Tekin~\etal~\cite{tekin2019h+} used the hand and object interaction as distinct features for the hand action recognition.
%
Another important factor in hand action estimation is eye gazing since people stare their hands and the target object while using their hands. 
Li~\etal~\cite{li2018eye} used eye gazing as a region of interest and extracted meaningful features in the area for solving hand action recognition.
%
Additionally, hand action recognition utilizes the sequence of the frames, and knowing the temporal relation between frames is critical for the perception of the action in the video.
Optical flow and motion vector are used to encode temporal information, and many works use these properties for hand action recognition~\cite{piergiovanni2019representation,zhang2016real,sun2018optical}. %
Aksoy~\etal~\cite{aksoy2010categorizing} used the semantic meaning between action and object relation for hand action recognition.
%

\textbf{Surface curvature in non-rigid object.}
Non-rigid objects such as hand have specific properties different from rigid objects.
For example, the diffusion of geometry-based descriptors constructs invariant metrics on rigid objects but these metrics are variant on non-rigid objects~\cite{wang2020diffusion}.
Therefore, surface information is used for shape features for the shape analysis of a non-rigid object.
Limberger~\etal~\cite{limberger2018curvature} proposed a shape descriptor based on a Lagrangian formulation of dynamics on the surface of the object for the retrieval task. 
The descriptor is a curvature-based scheme that identifies joints of the non-rigid object.  
Laskov~\etal~\cite{laskov2003curvature} proposed algorithms that use the relationship between Gaussian curvatures and geometric properties of a deformable non-rigid object for motion correspondence estimation accuracy.
Deboeverie~\etal~\cite{deboeverie2015curvature} segmented a human body based on the curvature of the human body given gray scales images, which showed that human body parts could be approximated by nearly cylindrical surfaces.
Chen~\etal~\cite{chen2020curvature} introduced the statistic index of curvatures from the curve in pixel space for estimating the human action recognition task. This work showed the effectiveness of the curvature representation as a spatiotemporal feature for high-level concept estimation.
Chang~\etal~\cite{chang2002hand} proposed a feature descriptor in curvature scale space with translation, scale, and rotation invariant properties for the recognition of hand pose for hand pose estimation.
As shown in the works above, curvature is a robust feature for a non-rigid object, such as a hand in our case.
In our work, we use the mean curvature of the primary hand for estimating hand actions.

\textbf{Hand grasp type for hand action.}
The grasp type is a symbolic representation of human intention while performing hand-related tasks.
Therefore, many works established the taxonomy of the grasp type for the manipulation applications in robotics and cognitive recognition in computer vision.
Feix~\etal~\cite{feix2015grasp} established the taxonomy of grasp type for one hand which is static and stable grasps. The taxonomy consists of 33 grasp types based on the opposition type, the virtual finger, the position of thumb, and the power adjustment of grasp based on the object shape. 
Cutkosky~\etal~\cite{cutkosky1989grasp} developed analytical models to  describe grasps in a manufacturing environment and established a taxonomy of manufacturing grasps which are used in designing robot hands.
Yang~\etal~\cite{yang2015grasp} used grasp types as a symbolic representation for reasoning human actions and used them as cognitive features for human intention prediction by applying the grasp types into the action segmentation task.
Cai~\etal~\cite{cai2016understanding} incorporated grasp type directly into hand action recognition task by learning the relation between grasp types and object attributes.
Underline assumption is that the grasp types and holding objects contain complementary information for recognizing the hand action.
We notice that these works only consider the grasp type but not the pre-grasp type: flat hand. 
We propose the concept of the pre-grasp type which is frequently appearing in hand action videos.

\section{Methods}

\subsection{Overview}
The purpose of our work is to develop a robust framework that estimates the hand action given videos which were taken from an egocentric-view RGB camera. 
For clarification, hand action covers~ \textit{verb~$+$~object} such as "drink with a mug" and also action without object such as \textit{verb} "high-five".
The framework predicts a hand action from a sequence of frames $\textbf{I}^t$ $\in$ $\mathbb{R}^{W\times H}$, where $\{0$ $\leq$ $t$ $\leq$ $N$, \ $N$ $\in$ $\mathbb{N}$ $\}$.
Here, we denote the cropped image from an input frame with a bounding box~(BBox) of the right hand and the object on the right hand as $h_r$ and $o$, respectively.
The framework first estimates $D(\textbf{I}^t) \eq \{h_r, o\}$ ,where $D$ is the object detector. 
With the $h_r$, the hand grasp type estimator, denoted as $f_g$, estimates hand grasp type $h_g\eq f_g(h_r)$ and successively synthesizes the mean curvature $H$ of the hand (see Fig.~\ref{fig:curv_net}).
Cropped hand and object images are fused by an encoder which generates a relation feature. 
Finally, a local feature is created by merging hand type grasp embedding, mean curvature vector, and interaction embedding.
To inform the global structure of the scene, a global feature is extracted from the object detector network. 
Local and global features are combined with series of fully-connected layers and used as a frame embedding~(see Fig.~\ref{fig:pipeline}), denoted as $FE$.
Finally, every $FE_t$ in a video sequence are batched and fed into the temporal model which is bi-directional Gated Recurrent Units (GRU) module for the final hand action recognition prediction.

\subsection{Hand and Object Detection}
The framework identifies the primary hand and target object, which defines a hand action. 
The primary hand is the right hand in the target dataset. Therefore, we define the right hand as the primary hand in our experiment setting.
We use an object detection network to identify the primary hand and target object in the scene.
To do so, we label the right/left hand and the target object, and we refine the object detection network with the labeled dataset.
Defining the target object does not require extra labeling effort since the action name itself contains the object name; we define the action named as a combination of verb and target object name: for example, "pouring the orange juice". 
Even though there is no object name in an action, picking one frame in the video is sufficient to identify the target object.
Additionally, it is much easier to annotate with a 2D BBox comparing with the 6D object pose annotation.
The right hand is determined among predicted hands by their relative location on the egocentric video, assuming there is only one person and the person is always using the right hand as a primary hand.
If only one hand is presented in the image, then that hand is considered as the primary hand.

The relation between the primary hand and target object are implicitly encoded with another deep neural network.
The output of the network is an interaction embedding in the framework and used for constructing a local feature. 
The interaction embedding represents the relation of hand and object, which is implicitly trained. 
Since the object detection network sees whole scene of the input frame, the intermediate feature vector is extracted as a global feature for creating a frame embedding.

The object detector utilized in the proposed framework is the YOLOv4~\cite{bochkovskiy2020yolov4} network architecture which was chosen due to its state-of-the-art performance. We tested and used three versions of YOLOv4 that differed from the original architecture through each model's depth and width multipliers (see Table~\ref{table:detector}). Depth refers to how deep the layers throughout the network are and how many times they are stacked on to themselves. Width refers to how wide the layers throughout the network are and how large the area of each layer is individually. We follow the training method from the original paper~\cite{bochkovskiy2020yolov4}.

\begin{table}[h]
  \begin{center}
    \begin{tabular}{l|r|c|c}
    \hline
     \multicolumn{1}{c|}{Model} &\multicolumn{1}{c|}{ Size (M)} &Width Multiplier&Depth Multiplier\\ 
    \hline
    YOLOv4-S&9.124&0.50&0.33 \\
    \hline
    YOLOv4-M&24.369&0.75&0.67 \\
    \hline
    YOLOv4-L&96.433&1.25&1.33\\
    \hline
    \end{tabular}
  \end{center}
  \caption{Configuration of three different variations of YOLOv4 for the primary/assistive object/hand detection.}
  \label{table:detector}
\end{table}

\subsection{Local Network}
\begin{figure}[h]
 \begin{center}
    \includegraphics[width=1.\linewidth]{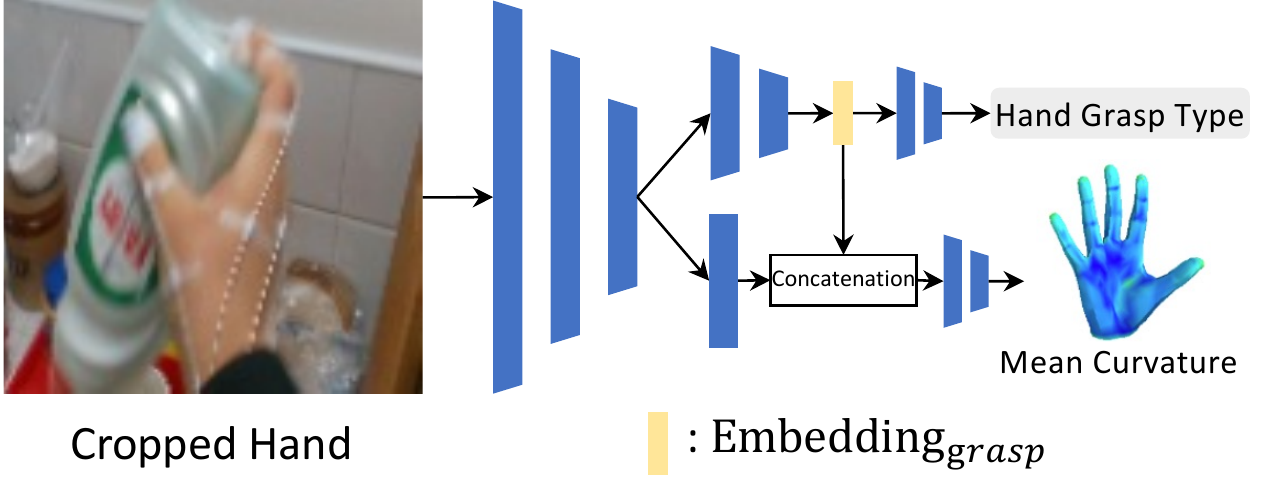}
 \end{center}
 \caption{The local network estimating a hand grasp type and mean curvature of the cropped hand.}
 \label{fig:curv_net}
\end{figure}

Local network consists of two networks: hand grasp type estimator and mean curvature estimator.
The local network starts with a pre-trained ResNet34~\cite{he2016deep} on ImageNet, and we use it as our backbone network that extracts image features~(see Fig.~\ref{fig:curv_net}).
The local network estimates a hand grasp type and mean curvature of the primary hand, simultaneously.

\begin{sidewaysfigure}[p]
 \begin{center}
    \includegraphics[width=1.\linewidth]{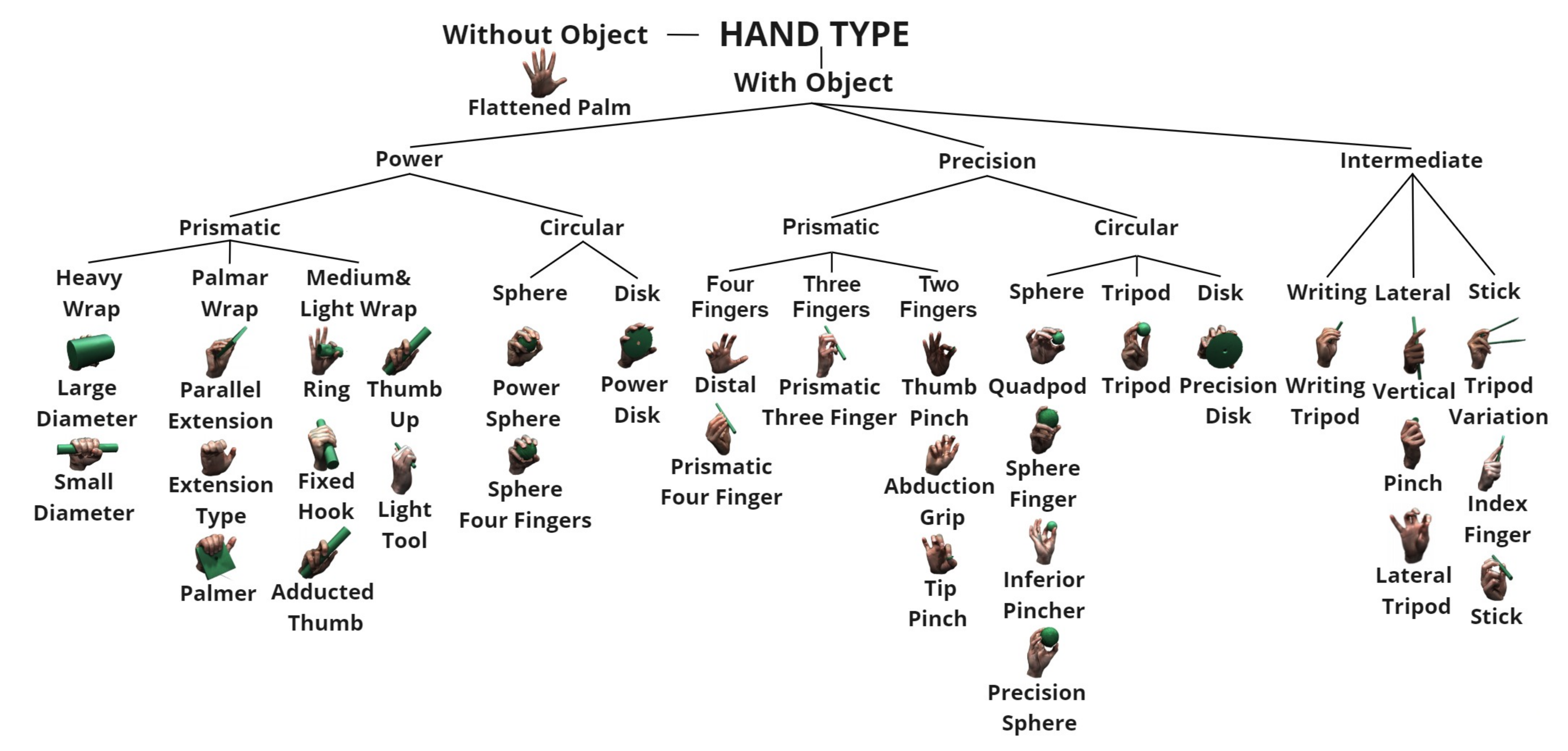}
 \end{center}
 \caption{Taxonomy of hand types for hand action recognition. Hand images are from MANO~\cite{romero2017embodied}.}
 \label{fig:hand_type}
\end{sidewaysfigure}

\textbf{Hand grasp type estimator.}
One challenging part of hand action recognition is occlusion of fingers from the object when holding an object and self-occlusion of the hand.
To relax this problem, we use the hand grasp type, which is a high-level concept of hand articulation.
For the hand grasp type estimation, we first establish the hand grasp type taxonomy~(see Fig.~\ref{fig:hand_type}). 
We construct the taxonomy of the hand grasp types from Feix~\etal~\cite{feix2015grasp} as our base taxonomy.
We notice that the hand actions of "high-five" and "receive a coin" do not require grasping. 
Therefore, our taxonomy includes not only hand grasp type with object but also grasp type without object and therefore covers more general cases of the hand grasp type: for example, "flattened palm" hand type in the taxonomy.
With this hand type taxonomy, we manually label the hand type in the video by marking the transition frames that show significant hand type changes.
Then we train the hand type estimator shown in Fig.~\ref{fig:curv_net} to estimate the hand type given a cropped image of the primary hand.

\textbf{Mean curvature estimator.}
Diverse hand actions consist of series of multiple hand types as shown in the scatter plot in Fig.~\ref{fig:scatter}.
As shown in the statistic results, hand grasp type is not a distinct factor to measure the detailed changes of hands for the hand action recognition task.
Therefore, we adopt hand surface modality that exposes local surface geometry of the hand.
Also, it is invariant from the global position of the hand and fingers.
To describe the hand surface, we use the mean curvature to represent the hand surface.
We use MANO~\cite{romero2017embodied} hand mesh model, which consists of 778 vertices and 1538 faces, and Baek~\etal ~\cite{baek2019pushing} provides fitted MANO hand model parameters on the dataset which we use in our experiment.
With the provided fitted hand mesh model, we compute the mean curvature of each vertex from 1-ring neighborhood of the vertex.
The mean curvature prediction model takes the primary hand cropped image as an input and jointly estimates a mean curvature and hand grasp type~(see Fig.~\ref{fig:curv_net}).

\subsection{Frame Embedding and Temporal Model}

\begin{figure}[h]
 \begin{center}
    \includegraphics[width=1.\linewidth]{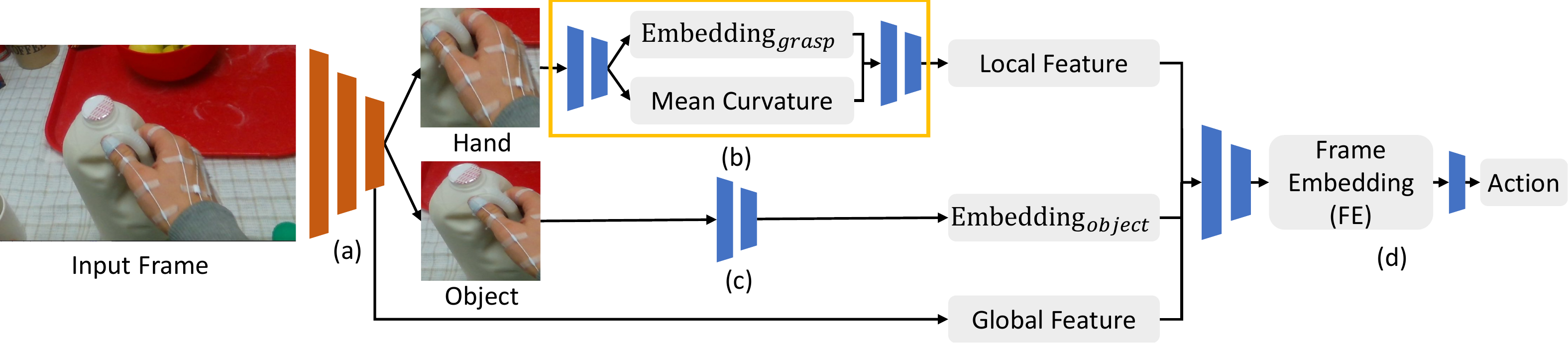}
 \end{center}
 \caption{Illustration of the frame embedding generator in the framework. The object detection network~(a) extracts a global feature and identifies the primary hand and target object from the input frame. The local network~(b) approximates mean curvatures of the primary hand and also generates a hand grasp type embedding. These two vectors are fused by series of fully-connected layers. Object network~(c) extracts an object embedding. The local feature, object embedding, and global feature are combined in the mixture network~(d) which generates a frame embedding.}
 \label{fig:pipeline}
\end{figure}

\textbf{Frame embedding generator.}
The frame embedding generator consists of four networks: object detection network, local feature network, object embedding network, and mixture network~(see Fig.~\ref{fig:pipeline}).
The generator identifies a primary hand and target object in the scene and crops them out. 
Then, the local network extracts a hand grasp embedding and mean curvature of the primary hand. 
Object network extracts an object embedding given the cropped target object image. 
Since the local feature and object embedding are captured from the cropped images, the features are not dependent on the global location in the scene.
Global information is further gained from the global feature which is obtained from the object detection network. 
These three embedding vectors are fused in the mixture network to generate a frame embedding.
Frame embedding contains not only global information but also isolated local information of both primary hand and target object.

\begin{figure}[h]
 \begin{center}
    \includegraphics[width=1.\linewidth]{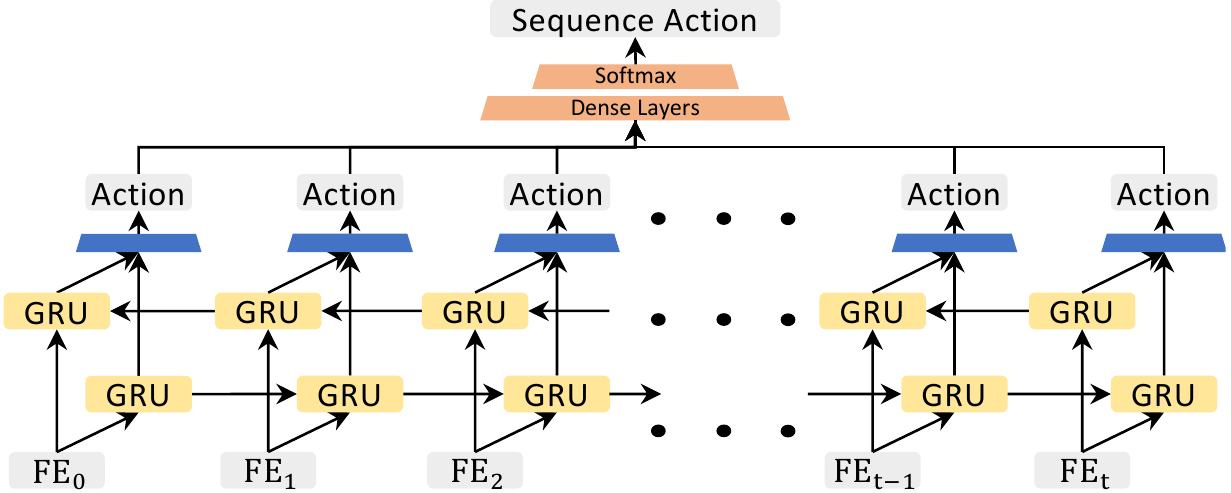}
 \end{center}
 \caption{Sequence action classification module for combining per-frame action estimation for final video action recognition of the hand with bi-directional GRU.}
 \label{fig:sequence}
\end{figure}

\textbf{Temporal model for action recognition.}
The hand action is not a static representation but a sequence of frames which contains spatiotemporal information.
Therefore, encoding the temporal information is essential for accurate hand action estimation.
In our framework, we use bi-directional Gated Recurrent Unites~(GRU) to capture the temporal relation in the video~(see Fig.~\ref{fig:sequence}). The output of each GRU unit estimates the action sequentially, and these action embedding are combined with three layers of fully-connected layers.
The fully-connected layers encapsulate the temporal information from each GRU unit to predict the final hand action in a video.

\subsection{Loss Function}
The notations in this section are as the following: $p^a, p^h,$ and $p^o$ are a probability of hand action, hand grasp type, and target object attribute, respectively. 
$C^a, C^h,$ and $C^o$ are a number of hand action classes, hand grasp type class, and target object attribute class, respectively. 
$\mathbf{v}\in \mathbb{R}^{778}$ is a vector that represents a mean curvature of the hand mesh model, and $N$ is the number of frames in the video.

\begin{flalign}
 & \mathcal{L}_{object} = \sum_{i}^{C^o} y_{i}^{o} log(p^{o}_{i})\label{eq:object}\\
 & \mathcal{L}_{local} \ \ = \sum_{i}^{C^h} y_{i} log(p^{h}_{i})  \, 
 + \alpha|| \hat{\mathbf{v}} - \mathbf{v} ||^{2}_{2} \label{eq:local} \\
 & \mathcal{L}_{action}^{f} = \sum_{i}^{C^a} y_{i}^{a} log(p^{a}_{i}) 
 + \beta\, \mathcal{L}_{object} \, 
 + \kappa\, \mathcal{L}_{local}\label{eq:action_f} \\
 & \mathcal{L}_{action}^{t} = \sum_{i}^{N} y_{i}^{a} log(p^{a}_{i}) \label{eq:action_t}
\end{flalign}

\ \ \ ,where $\alpha$$\eq$0.3, $\beta$$\eq$0.2, and $\kappa\eq$0.5. 

The framework consists of two major parts: frame embedding generator and temporal model. 
Also, the frame embedding generator consists of four networks: object detection network, local network, object network, and mixture network. 
The object detection network is trained separately with the loss function from YOLO~V4~\cite{bochkovskiy2020yolov4}. 
The local network and object network are also optimized separately by minimizing Eq.~\ref{eq:local} and Eq.~\ref{eq:object}, respectively. 
And then, the object detection network, local network, and object network are optimized, and the mixture network is added on top of these three networks. 
Then these networks are refined by minimizing Eq.~\ref{eq:action_f}.
After the frame generator is optimized, we freeze the network's parameters and jointly train the temporal model with Eq.~\ref{eq:action_t} by following the training methods in \cite{chapelle2010gradient,yi2016lift}.

\begin{sidewaysfigure}[p]
 \begin{center}
    \includegraphics[width=1.\linewidth]{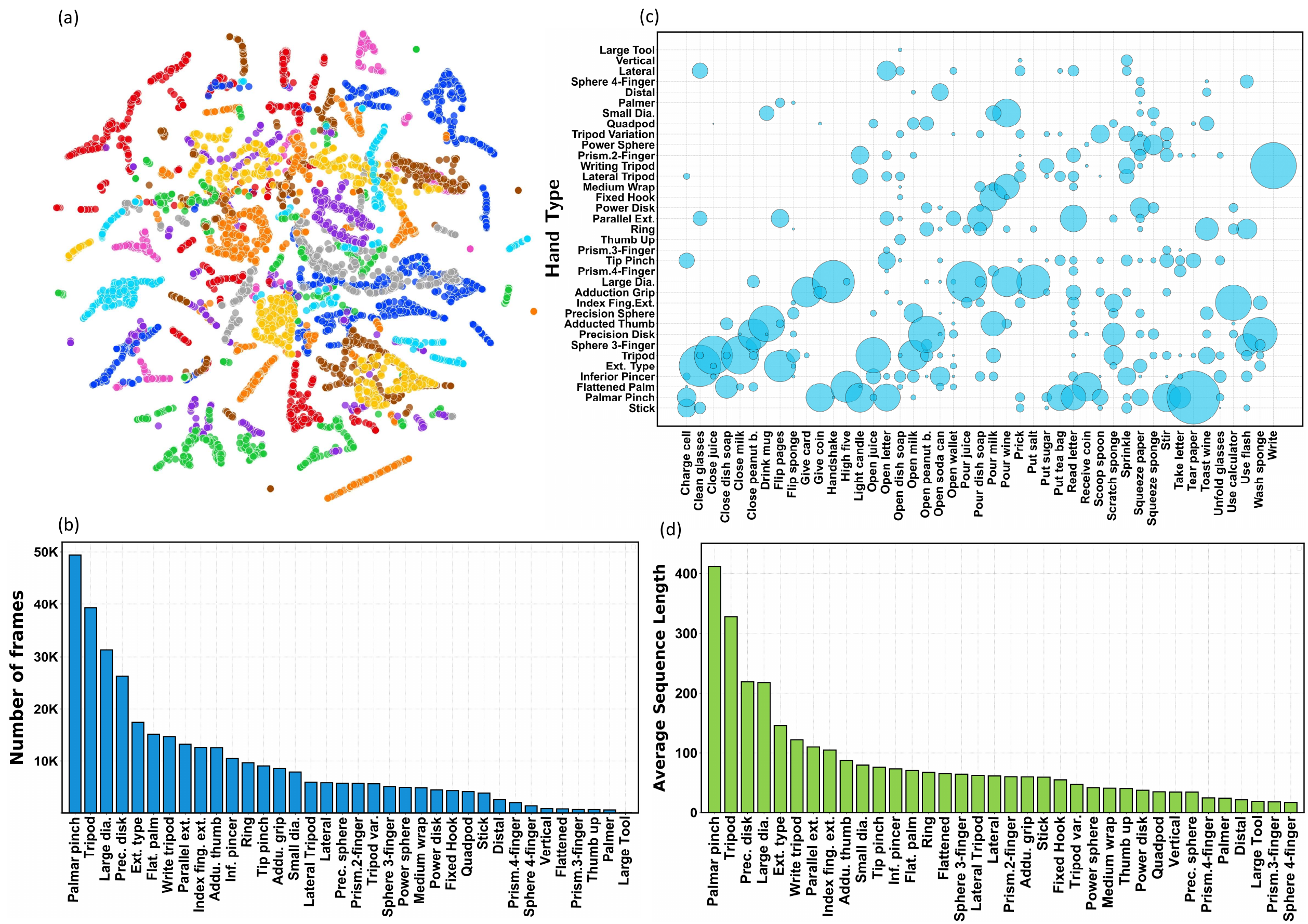}
 \end{center}
 \caption{(a) t-SNE visualization of action class embedding over dataset. (b) Number of frames per hand type. (c) Scatter plot showing the distribution of the hand type per hand action in the dataset. The y axis shows the hand type, and the x axis represents the action type. The size of the circles represents the frequency of occurrence of a hand type per action class. (d) Average number of frames in each video per hand type.}
 \label{fig:scatter}
\end{sidewaysfigure}

\section{Experiment}
In this section, we first elaborate on the dataset for the experiment.
Then, we illustrate an ablation study of each proposed method and compare our method with other state-of-the-art methods.

\subsection{Dataset}
We use the First-Person Hand Action (FPHA) dataset~\cite{garcia2018first} since this dataset is the only public dataset in 3D hand-object action recognition task with 3D-fitted MANO hand model.
FPHA consists of 1175 videos performed by 6 subjects, 24 different objects, and 45 different hand action categories.
The total number of frames is 105,459 with accurate 3D hand poses and 6D pose object annotations.
The data has a subset split named TinyFPHA that has 3D object mesh models which belong to 10 actions. 
Additionally, we manually annotate hand types in the videos by indexing the frame number if there are hand type transitions based on the hand grasp type taxonomy with 36 different hand types~(see Fig.~\ref{fig:hand_type}).
The distribution of the hand type of the dataset is shown in two histogram plots in Fig.~\ref{fig:scatter}.
From the histogram, we notice that the hand type is not evenly distributed.
We further analyze the dataset by plotting the distribution of hand type for each action class as a scatter plot~(see Fig.~\ref{fig:scatter}). 
From the scatter plot, we find that one action has multiple hand types.
Therefore, relying solely on the hand type for hand action estimation is not a good design choice for the pipeline.

\subsection{Experimental Results}
All the experiments in this paper use the dataset provided by FPHA~\cite{garcia2018first}.
The dataset is divided into train and test sets based on the split schema from the dataset.

\subsubsection{Object detection network and object network.}
For the primary hand and target object detection, we utilize the large YOLOv4~\cite{bochkovskiy2020yolov4} network that uses Mish activation~(YOLOv4-L). YOLOv4-L has 1.33 times more depth and 1.25 times more width to its layers than the original YOLOv4 layers. The network was trained on images with size 480 $\times$ 270 and a batch size of 8 for 20 epochs. For the evaluation, we use the Generalized Intersection over Union (gIoU) with a 0.05 gIoU loss gain and the mean average precision (mAP) at both IoU thresholds 0.5 and 0.5 to 0.95 as our performance metrics. The mAP of the YOLOv4-L is 78.12\% at the IoU threshold of 0.5 and 46.15\% at the IoU threshold of 0.5 to 0.95. Table~\ref{table:detector} describes the mAP values and the gIoU for the other versions of YOLOv4 with the same training conditions mentioned previously. The difference in size can be described based on a depth and width multiplier that is applied to every layer in the network, and these scales are relative to the original YOLOv4 layer sizes. Depth refers to how deep the layers throughout the network are and how many times they are stacked on to themselves. Width refers to how wide the layers throughout the network are and how large the area of each layer is individually. Each type of layer is represented by height $\times$ width $\times$ depth when defining the network architecture.   

For object network, we use ResNet18~\cite{he2016deep} pre-trained on ImageNet as the backbone followed by a 36-way fully-connected layer. We use 0.0001 as the learning rate and a batch size of 32. The network is fully trained within around 30 epochs. The network achieves accuracy of an accuracy of 96.42\% on the test set. After refining with the local network and mixture network, the network accuracy improves as 98.52\%.

\begin{table}[h]
  \begin{center}
    \begin{tabular}{l|r|c|c|c}
    \hline
     \multicolumn{1}{c|}{Model} &\multicolumn{1}{c|}{ Size (M)} &mAP@0.5 (\%)&mAP@0.5:0.95 (\%)&gIoU\\ 
    \hline
    YOLOv4-S&9.124&76.87&42.88& 0.037 \\
    \hline
    YOLOv4-M&24.369&77.22&45.00&0.036 \\
    \hline
    YOLOv4-L&96.433&78.12&46.15&0.035\\
    \hline
    \end{tabular}
  \end{center}
  \caption{Performance of the three different variations of YOLOv4 for the primary/assistive object and hand detection.}
  \label{table:detector}
\end{table}

\subsubsection{Ablation studies and implementation details.}
We perform ablation studies to evaluate our proposed methods. 
There are two major parts on the framework: frame embedding generator and temporal model.
The feature generator consists of three networks: local network, global network, and mixture network.
The local network captures local feature of hand, and the global network captures scenery information.
These two feature vectors are fused by the mixture network.
To evaluate each method, we conduct the experiment on TinyFPHA.
For implementation details, the frame embedding generator uses YOLOv4-L for detecting the primary hand and target object.
We use ResNet18 for the backbone network which is pre-trained on ImageNet in "local network" and "object network" as shown in Fig.~\ref{fig:pipeline}.
The object network takes a cropped image of target object image with a size of $224\times224$ and outputs a vector size of 256.
Similarly, the local network takes a cropped image of primary hand image of size 224$\times$224. 
The network first encodes the image into a feature vector with the pre-trained backbones as a feature extractor~(see Fig.~\ref{fig:pipeline}).
In the local network, this feature vector from the backbone is used to predict the hand grasp type by solving the classification problem.
The intermediate feature vector of the hand type network is concatenated with the feature vector from the backbone for estimating the hand curvature with series of FC layers~(see~Fig.~\ref{fig:curv_net}). 
For training details, we use 0.0004 as the running rate and 64 as the batch size.
The network is fully trained with 200 epochs. 
$L_2$ loss of the test set of the mean curvature is 3.238, and the hand grasp type accuracy is 97.32\% after refining the network with the object network and the mixture network.
The accuracy of hand type is not as high as the hand action accuracy.
This is because the distribution of hand type in the dataset has a long tail as shown on the histogram in Fig.~\ref{fig:scatter}.
We tried four different types of curvature: mean curvature, Gaussian curvature, maximum curvature, and minimum curvature for the framework. 
Using the mean curvature performs the best as shown in Table~\ref{table:ab}.
The Gaussian curvature becomes negative value whenever principle curvatures are not in the same direction, which makes hard to find optimal solution of the network.
From our experimental results, we observe that adding the mean curvature creates a performance leap on both datasets~(see~Tables~\ref{table:tiny}~and~\ref{table:full}).
This is explainable as shown in our t-SNE plot~\cite{maaten2008visualizing} of primary hand mean curvature in Fig.~\ref{fig:scatter}, which shows a tight clustering group by action.
We also notice that using the hand grasp type estimation helps improving the hand action recognition task~(see~Tables~\ref{table:tiny}~and~\ref{table:full}).
The distribution of hand grasp type is unique per hand action (see the scatter plot (c) in Fig.~\ref{fig:scatter}), implying that hand grasp type is a distinct feature that can be used as prior knowledge for hand action recognition estimator.
For the temporal model, we use a 2-layer bi-directional GRU with a hidden state size of 256.
To prevent overfitting, we augment data by randomly changing the saturation, hue, and exposure of the images by a maximum factor of 50\% and randomly translate and rotate them by a maximum offset of 10\%.
For optimizing the network, we use a stochastic gradient descent optimization method with 0.001 learning rate and decay by half on every 50 epochs. The action estimator is trained with a batch size of 64 for 200 epochs.
Having temporal model in the framework brings the temporal information to the framework, resulting in performance improvement~(see Tables~\ref{table:tiny}~and~\ref{table:full}).
The action recognition is not a frame estimation, but it involves temporal relation between frames, which explains the performance improvement. 

\begin{table}[h]
  \begin{center}
    \begin{tabular}{l|c}
    \hline
    \multicolumn{1}{c|}{Method} &  Accuracy (\%)\\ 
    \hline
    Gaussian Curvature & 96.52\\
    Maximum Curvature & 95.97 \\
    Minimum Curvature & 95.35 \\
    Mean Curvature & \textbf{97.54}\\
    \hline
    \end{tabular}
  \end{center}
  \caption{Evaluation of four different curvatures for hand action recognition on TinyFPHA.}
  \label{table:ab}
\end{table}

\subsubsection{Comparison with other methods.}
\label{sec:ex_ha}
We compare our method with other state-of-the-art works in two datasets: TinyFPHA and FPHA.
Many methods~\cite{tekin2019h+,garcia2018first,li2021trear} used 3D location of hand pose and 6D object pose explicitly to capture the object hand relation information. 
These pose information helps locating the position of fingers on the object in the scene.  
From the experimental results, using hand pose and object pose improves the performance of the action estimation.
We find that hand grasp type estimation improves the accuracy by 7.7\% and adding mean curvature feature enhances the accuracy by 1.47 \% in TinyFPHA dataset.
Adding the temporal model increases extra 0.72\% improvement in hand action accuracy. 
These results shows that hand grasp type and mean curvature feature are critical in hand action recognition.
Depth information was widely used in the hand action recognition task~\cite{garcia2018first,ohn2014hand,hu2015jointly,oreifej2013hon4d,rahmani20163d,li2021trear} by using depth camera to estimate hand action recognition.
The drawback of using depth camera is not robust on outside situation, which makes huge performance gap between indoor and outdoor situation.
In full FPHA dataset, adding mean curvature improves the accuracy by 2.37\%.
Adding the temporal model enhances the performance by 1.9\% in full FPHA dataset.
The experimental results shows that the improvement gap becomes larger when the size of the dataset gets bigger.
Our method shows the accuracy of 95.62\% on the full FPHA dataset and 97.54\% on the TinyFPHA dataset, outperforming the other methods by a large margin.
Hand grasp type, mean curvature of the primary hand mesh, and temporal model contribute significantly to the performance in both datasets, and our method outperforms the state-of-the-art methods by a large margin (see Tables~\ref{table:tiny}~and~\ref{table:full}).

\begin{table}[h]
  \begin{center}
    \begin{tabular}{c|l|c}
    \hline
    \multicolumn{1}{c|}{Method} & \multicolumn{1}{c|}{Model} & Accuracy (\%)\\ 
    \hline
    & OP &87.45\\
    FPHA~\cite{garcia2018first}& 3D HP &74.45\\
    & HP + OP &91.97\\
    \hline
    &Image &85.56\\
    &HP &89.47\\
    H+O~\cite{tekin2019h+}&OP &85.71\\
    &HP + OP&94.73\\
    &HP + OP + Interaction &96.99\\
    \hline
    &Image & 87.58\\
    &Grasp Type~(GT) & 95.35\\
    \multirow{2}{0.75cm}[0.65cm]{Ours} 
    &GT + Mean Curvature~(MC) & 96.82\\
    &GT + MC + Temporal Model & \textbf{97.54}\\
    \hline
    \end{tabular}
  \end{center}
  \caption{Performance evaluation of different methods on TinyFPHA for the hand action recognition. 
  HP, OP, GT, MC, and TM stand for hand pose, object pose, grasp type, mean curvature, and temporal model, respectively.
  Our method outperforms other methods by a large margin.}
  \label{table:tiny}
\end{table}

\begin{table}[h]
  \begin{center}
    \begin{tabular}{l|c|c}
    \hline
    \multicolumn{1}{c|}{Method} & Input modality & Accuracy (\%)\\ 
    \hline
    FPHA~\cite{garcia2018first}& Depth&72.06\\
    HOG~\cite{ohn2014hand}   ,&Depth & 59.83\\
    JOULE~\cite{hu2015jointly} &Depth & 60.17\\
    NON4D~\cite{oreifej2013hon4d} &Depth & 70.61\\
    Novel View~\cite{rahmani20163d}&Depth & 69.21\\
    Trear~\cite{li2021trear}&Depth & 92.17\\
    \hline
    H+O~\cite{tekin2019h+}&RGB &82.43\\
    JOULE~\cite{hu2015jointly} &RGB & 66.78\\
    Two stream - all~\cite{feichtenhofer2016convolutional} & RGB &75.30\\
    Two stream - RGB~\cite{feichtenhofer2016convolutional} & RGB &61.56\\
    Two stream - flow~\cite{feichtenhofer2016convolutional} & RGB &69.91\\
    \hline
    Ours - Grasp Type~(GT) & RGB & 91.35\\
    Ours - GT + Mean Curvature~(MC)& RGB & 93.72\\
    Ours - GT + MC + Temporal Model & RGB & \textbf{95.62}\\
    \hline
    \end{tabular}
  \end{center}
  \caption{Performance evaluation of different methods on FPHA for the hand action recognition. 
  Abbreviations are the same as in Table~\ref{table:tiny}.
  Our method outperforms other methods by a large margin.}
  \label{table:full}
\end{table}

\section{Conclusion and Future Work}
In this work, we tackle a hand action recognition problem with an egocentric-view video taken from a RGB camera.
We propose a method that does not require neither 6D object pose nor a depth sensor for estimating hand action recognition.
Instead, we use hand grasp type and mean curvature to capture the local feature of high-level concept of hands which is invariant from global position of the hand location. 
Additionally, the framework estimates the target object attributes and combines the local feature with global scene information to integrate the object-hand relation. 
These priors improve the hand action recognition accuracy by a large margin.
Finally, the temporal model captures spatiotemporal information to estimate the hand action in the video.
The proposed method outperforms the state-of-the-art methods.
As future works, the proposed method can be applied to more general cases that involve multi-person in the view.
Also, the work can be applied to 3D hand pose estimation while interacting with objects.
\label{sec:ex}

\section{Acknowledgement}
This work was supported by Institute of Information \& Communications Technology Planning \& Evaluation~(IITP) of the Korean government under grants No.~2019-0-00079~(Artificial Intelligence Graduate School Program of Korea University) and No.~2017-0-00417~(Openholo Library Technology Development for Digital Holographic Contents and Simulation) and U.S. National Science Foundation grant Future of Work at the Human Technology Frontier
FW-HTF 1839971 (http://www.nsf.org/). We also acknowledge the support of the Feddersen Chair Funds to Professor Ramani.
Any opinions, findings, and conclusions or recommendations expressed in this material are those of the authors and do not necessarily reflect the views of the funding agency.

\clearpage


\bibliography{egbib.bib}

\end{document}